\documentclass[10pt,journal,compsoc]{IEEEtran}

\ifCLASSOPTIONcompsoc
  \usepackage[nocompress]{cite}
\else
  \usepackage{cite}
\fi
\usepackage{graphicx}
\usepackage{amsmath}
\DeclareMathOperator*{\argmax}{arg\,max}

\usepackage{amssymb}
\usepackage{multirow, booktabs}
\usepackage{xspace}
\usepackage{makecell}
\usepackage{verbatim}
\usepackage{xcolor}
\usepackage[capitalize]{cleveref}
\crefname{section}{Sec.}{Secs.}
\Crefname{section}{Section}{Sections}
\Crefname{table}{Table}{Tables}
\crefname{table}{Tab.}{Tabs.}
\ifCLASSINFOpdf
\else
\fi
\begin{document}
\definecolor{mygray}{gray}{0.4}
%
\title{What Makes for Good Tokenizers \\in Vision Transformer?}

\author{Shengju Qian,
        Yi Zhu,
        Wenbo Li,
        Mu Li,
        Jiaya Jia,~\IEEEmembership{Fellow,~IEEE}
\IEEEcompsocitemizethanks{\IEEEcompsocthanksitem S. Qian, W. Li, and J. Jia are with the Department
of Computer Science and Engineering, The Chinese University of Hong Kong.\protect\\
E-mail: sjqian@cse.cuhk.edu.hk
\IEEEcompsocthanksitem Y. Zhu and M. Li are with Amazon AI.}
\thanks{Manuscript received April 19, 2005; revised August 26, 2015.}}

%
%

\markboth{Journal of \LaTeX\ Class Files,~Vol.~14, No.~8, August~2015}%
{Shell \MakeLowercase{\textit{et al.}}: Bare Demo of IEEEtran.cls for Computer Society Journals}
%



\IEEEtitleabstractindextext{%
\begin{abstract}
The architecture of transformers, which recently witness booming applications in vision tasks, has pivoted against the widespread convolutional paradigm. Relying on the tokenization process that splits inputs into multiple tokens, transformers are capable of extracting their pairwise relationships using self-attention. While being the stemming building block of transformers, what makes for a good tokenizer has not been well understood in computer vision. In this work, we investigate this uncharted problem from an information trade-off perspective. In addition to unifying and understanding existing structural modifications, our derivation leads to better design strategies for vision tokenizers. The proposed Modulation across Tokens~(MoTo) incorporates inter-token modeling capability through normalization. Furthermore, a regularization objective TokenProp is embraced in the standard training regime. Through extensive experiments on various transformer architectures, we observe both improved performance and intriguing properties of these two plug-and-play designs with negligible computational overhead. These observations further indicate the importance of the commonly-omitted designs of tokenizers in vision transformer.
\end{abstract}

}

\maketitle

\IEEEdisplaynontitleabstractindextext

%
\IEEEpeerreviewmaketitle

\IEEEraisesectionheading{\section{Introduction}\label{sec:intro}}
Serving as a prevalent model in natural language processing~(NLP), advances in transformers have driven progress in computer vision. With a great leap made by Vision Transformer~(ViT)~\cite{dosovitskiy2021an}, we have witnessed an increasing effort on adapting this dominating language paradigm to vision. Meanwhile, performance on many downstream tasks such as video recognition~\cite{zhang2021vidtr,shao2020temporal},  object detection~\cite{carion2020end,zhu2020deformable}, and semantic segmentation~\cite{SETR,xie2021segformer,EDT} have upgraded considerably, suggesting the potential of transformer as a primary backbone for vision applications.

While convolution excels at capturing local interactions with its inherent inductive bias such as translation invariance and local connectivity, self-attention in transformers introduces appealing advantages of long-range context modeling and parameter efficiency. In contrast to the feature extraction pipeline in convolutional neural networks~(CNNs), we identify the forward process of transformers as two separate stages: the tokenization head that splits inputs into multiple tokens, and the follow-up transformer body that models the pair-wise correlations among the obtained tokens. This reliance on specific tokenization method possibly introduces a bottleneck into both language and vision transformers that limits their capabilities, as discussed in~\cite{tay2021charformer,el-boukkouri-etal-2020-characterbert,xiao2021early}. 

Being a fundamental prompt in NLP, tokenization strategies have evolved rapidly from the standard rigid tokenization~\cite{sennrich-etal-2016-neural}. To cope with variation in language, probabilistic segmentation algorithms like subword regularization~\cite{kudo2018subword} have been proposed. Character-level tokenizations lately emerge as a powerful solution in dealing with languages without whitespace separation~\cite{clark2021canine,tay2021charformer}.
Compared with language sequences, natural images are more complex and have no concise grammar. However, the naive patchify strategy adopted in ViT splits images into non-overlapping patches during tokenization, which are then fed into transformer blocks. Compared to the more continuous and natural distribution of pixels, tokenization and self-attention performed on the discontinuous token embeddings are coarse-grained and may hinder the modeling power of transformers. Recent works alternatively exploited a convolutional stem~\cite{xiao2021early,wang2021pyramid} or an overlapping patch embedding~\cite{wu2021cvt,yuan2021tokens} to replace the naive tokenization in ViT~\cite{dosovitskiy2021an}. While existing works commonly offer system-level development, the influence from different tokenizer designs still lacks principled discussion and analysis. Meanwhile, the training instability and data-inefficiency still exist as major curse for vision transformer. As discussed in recent studies~\cite{xiao2021early,chen2021empirical}, the naive tokenization stem possibly accounts for these drawbacks.

In this work, we provide an alternative perspective on designing good vision tokenizers. We first conjecture that a good vision tokenizer serves as the information bridge for follow-up transformer blocks, while connects the representations from input images to split tokens. Then we demonstrate that existing structural modifications fit in this formulation. In addition to offering and defending this holistic understanding that bridges different designs of tokenizers, we further exploit the aforementioned assumption, and conduct in-depth analysis of a good vision tokenizer towards its better normalization and optimization objectives. To summarize, our contributions are:


\begin{itemize}
    \item We investigate the vision tokenizer design from a information trade-off perspective, and analyze existing design choices from this viewpoint. We also demonstrate that unlike language, naively increasing mutual diversity across tokens doesn't guarantee better performance.
    \item We tailor a novel regional normalization strategy for tokenizer, which models better inter-token and intra-token interactions in input images.
    \item Inspired by the findings, we further incorporate an additional objective that regularizes the original optimization process of tokenizers, which leads to better performance and adaptability to simpler training recipes.
    \item We evaluate our strategy on various transformer families. Experiments show that our simple yet effective design boosts sophisticated architectures with negligible overhead. The comparisons also demonstrate that vision tokenizer, while lacks in-depth analysis, should be featured prominently in the transformer pipeline. 
\end{itemize}

\section{Related Works}
\label{sec:related}
\subsection{Vision Transformers.} 

The transformer architecture was introduced in machine translation~\cite{vaswani2017attention}, and gradually became a primary model for various NLP tasks. The ViT~\cite{dosovitskiy2021an} structure closes the gap with CNN models on image recognition, and further demonstrates data efficiency in DeiT~\cite{touvron2020training} with a distillation token and stronger augmentations~\cite{cubuk2020randaugment,cubuk2018autoaugment}. In order to obtain suitable architectures that generalize better to vision, researchers have made progress that introduces more 
hand-crafted designs~\cite{chen2021visformer,chen2021crossvit,wu2021cvt,yuan2021tokens}. Recent structural modifications include improving the positional encoding as in~\cite{chu2021conditional,xu2021coscale,dai2021coatnet}, reducing the computational burden via either a local self-attention~\cite{liu2021swin,vaswani2021scaling} or a refined global self-attention~\cite{wang2021pyramid,yang2021focal,SOFT,chu2021Twins}. Meanwhile, the feature redundancy in vision transformers is observed, which leads to diversified strategies like adaptive length~\cite{wang2021not} and redundancy reduction~\cite{rao2021dynamicvit,pan2021ia}. 

\subsection{Tokenization in Language and Vision.} 
Tokenization serves as a stemming stage in transformers, for both language and vision. Being a long-standing linguistics problem, recent tokenizers such as FRAGE~\cite{gong2018frage} and CharacterBERT~\cite{el-boukkouri-etal-2020-characterbert,tay2021charformer} have demonstrated superiority in deep transformers when modeling languages over the widespread BPE~\cite{sennrich-etal-2016-neural} and WordPiece~\cite{wu2016googles} . The heavy reliance on proper tokenizers for vision transformers has also been observed in the training instability of MoCov3~\cite{chen2021empirical} and ViT stem~\cite{xiao2021early}. Further evidence appears in the redundancy among visual tokens~\cite{pan2021ia,rao2021dynamicvit,marin2021token}. Therefore, a convolutional stem~\cite{xiao2021early} and overlapped token embeddings~\cite{yuan2021tokens,chen2021crossvit,wang2021pvtv2} are explored to replace the naive patchify tokenization. Meanwhile, PnP-DETR~\cite{wang2021pnp}, TokenLearner~\cite{ryoo2021tokenlearner}, PS-ViT~\cite{psvit}, and Token Labeling~\cite{jiang2021all} also explore sampling and learning strategies. While these system-level designs are achieving better downstream performance and suggesting the importance of tokens, there lacks insightful discussions beyond structures for vision tokenizers.

\section{Preliminary\label{sec:theory}}

\subsection{Mutual Information between Random Variables}

Mutual information initially measures dependencies between random variables. Given $\mathbf{A}$ and $\mathbf{B}$, $I(\mathbf{A};\mathbf{B})$ estimates the ``amount of information'' learned from $\mathbf{B}$ about the other variable $\mathbf{A}$ and vice versa. The mutual information can be formally defined as:

\begin{equation}
I(\mathbf{A};\mathbf{B}) = H(\mathbf{A}) - H(\mathbf{A}|\mathbf{B}) = H(\mathbf{B}) - H(\mathbf{B}|\mathbf{A})
\label{eq:mutual-info}
\end{equation}

Let $\mathbf{A} \in \mathcal{R}^{h \times w \times 3}$, $\mathbf{B} \in \mathcal{R}^{n_\text{token} \times l_\text{token}}$ denote an input image and its partitioned tokens, where $n_\text{token}$ and $l_\text{token}$ respectively refer to the number and dimension of tokens.

\begin{equation}
\begin{aligned}
    I(\mathbf{A};\mathbf{B}) & = H(\mathbf{A}) -H(\mathbf{A}|\mathbf{B}) \\
                    & \geq H(\mathbf{A}) - \mathbb{R}(\mathbf{A}|\mathbf{B})
    \end{aligned}
\label{eq:mutual-rec}
\end{equation}
where $\mathbb{R}(\mathbf{A}|\mathbf{B})$ denotes the expected error for reconstructing $\mathbf{A}$ from $\mathbf{B}$.
$H(\mathbf{A})$ refers to $\mathbf{A}$'s marginal entropy, which is treated as a constant in this formulation~\cite{vincent2010stacked,kingma2013auto,hjelm2018learning}.

In this work, we strive to understand whether vision tokenization fits in this information bottleneck~\cite{michael2018on, qian2019make,qian2019aggregation} principle, which disentangles vision transformers into two distinct phases of representation learning. 

\subsection{Estimating Potential Bottleneck in Tokenizers with Conditional Entropy} 

For vision transformer, the tokenizer divides input image to multiple patch tokens and the cascaded transformer blocks perform global context modeling across the tokens. Compare to the ``continuous'' representation of image, self-attention performed can be regarded as subsampled operations on its discontinuous token embeddings. Analogous to linguistic analysis~\cite{voita-etal-2019-bottom}, we attempt to testify whether vision tokenizer poses an information bottleneck.

\begin{figure*}[htb]
\centering 
\includegraphics[width=0.9\textwidth]{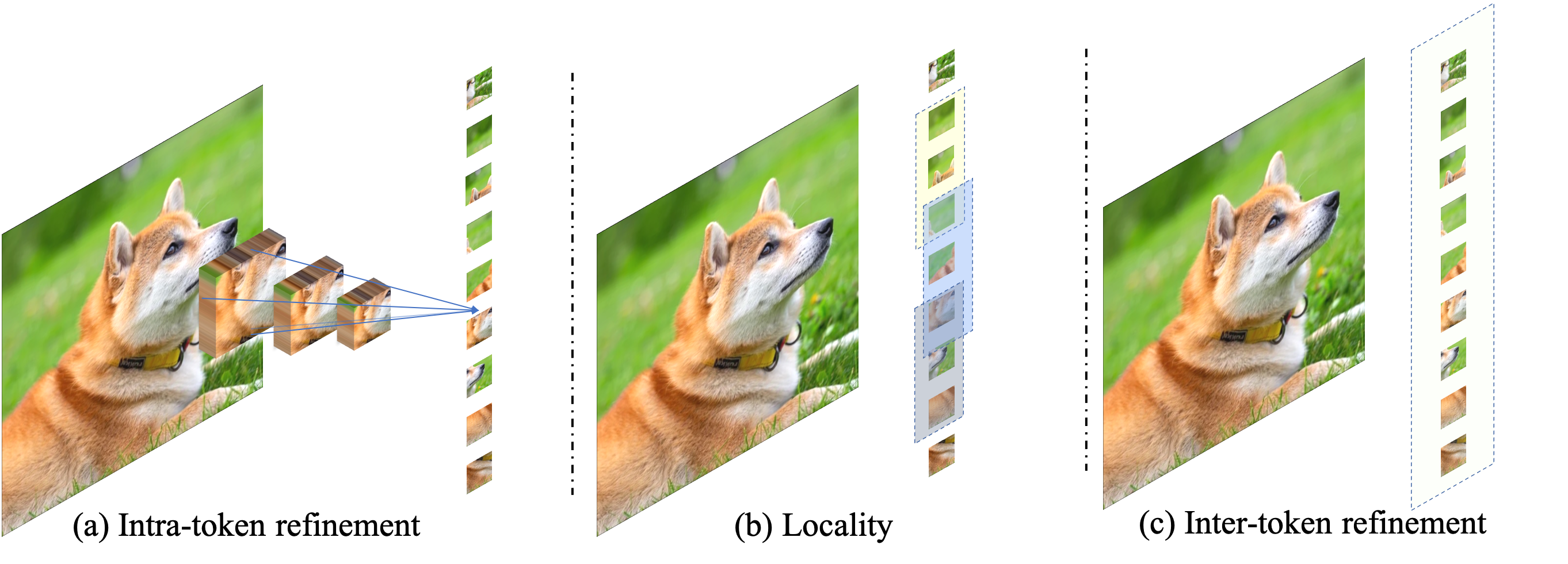}
\caption{\textbf{Illustration on three major aspects of structural improvements for vision tokenizers.} Note that we make these modifications to exemplify our perspective, which is not designated for superior performance or novelty. For intra-token refinement, we adopt the multi-scale feature extraction as in~\cite{wang2021crossformer,Fan_2021_ICCV}. We utilize the overlapping embedding as in~\cite{yuan2021tokens,wang2021pvtv2} and self-attention to represent locality and intra-token refinements. The tokens in dotted regions with the same color perform interactions. Locality could be viewed as a local case of Inter-token refinement, where Inter-token refinement emphasize more on global modeling. }
\label{fig:structure}
\end{figure*}

Different from the mathematical definition of MI, the bottleneck in vision tokenizers denotes the post-tokenizer information accessibility. Empirically, we characterize this bottleneck as the information lost in tokenization. Therefore, we estimate $I(\mathbf{A};\mathbf{B})$ by a parametric decoder that obtains the minimal $\mathbb{R}(\mathbf{A}|\mathbf{B})$ between images and their respective tokens. In practice, we adopt a simple three-layer decoder and optimize it using $L_2$ loss towards the inputs following~\cite{hjelm2018learning,belghazi2018mutual}. The final reconstruction error reflects the empirical conditional entropy and information accessibility between the input image and its split tokens.

\section{\label{sec:structure}Structural designs for Vision Tokenizer}

Due to the quadratic complexity of self-attention operation, naive extension of ``per-word'' tokenization in NLP doesn't scale to image pixels. A widespread practice is to process each input image into a sequence of non-overlapping patches as in ViT~\cite{dosovitskiy2021an}. Despite alleviating computational costs, the ``patchify'' operation causes other problems such as training instability~\cite{chen2021empirical,xiao2021early} and aliasing~\cite{qian2021blending}. To mitigate the drawbacks, architectural modifications have been made towards more suitable vision tokenizers~\cite{Fan_2021_ICCV,xiao2021early,yuan2021tokens}. To facilitate understanding, we propose to summarize existing structure-level improvements to three major aspects:

\begin{enumerate}
    \item \textbf{Intra-token refinement:} As implemented by a stride-$p$, $p \times p$ convolution, naive ``patchifying'' fails to capture rich context inside the tokens. It's straightforward to assign the tokenization encoder with stronger capability of feature extraction. Recent progress that exploited convolutional stem~\cite{xiao2021early,wang2021pyramid}, multi-scale embeddings~\cite{chen2021crossvit,wang2021crossformer,Fan_2021_ICCV}, or an additional transformer that models sub-token~\cite{han2021transformer} can be classified in this category. 
    \item \textbf{Locality:} The partition strategy splits the continuous image content into relatively divided portions. As pixel statistics of nearby tokens indicate better positions and connectivity, boosting patch tokens with locality also helps. Overlapping~\cite{yuan2021tokens,wang2021pvtv2} and uneven~\cite{chen2021dpt} token embeddings have been tailored.
    \item \textbf{Inter-token refinement:} Due to the semantic complexity, similar instances in different images might require distinctive features. Modeling inter-token relationship intuitively helps tokenizers encode better features. This refinement suggests performing global context modeling inside the tokenizers, which has demonstrated its effectiveness in learning~\cite{ryoo2021tokenlearner,jiang2021all,yuan2021tokens} or sampling~\cite{psvit,wang2021pnp,marin2021token} strategies.
\end{enumerate}

While diverse structural refinements have been explored in recent works, MoCov3~\cite{chen2021empirical} empirically shows that naive frozen tokenization enhances contrastive training stability. As such randomly-initialized projection possibly maximize the information accessibility, it's crucial to validate the existence of potential bottleneck in vision tasks. 

\subsection{Comparison across Potential Structural Designs}

In order to study the corresponding influence, we adopt four modifications from existing transformer frameworks to the original ViT tokenizer, which includes the three categories illustrated in Figure~\ref{fig:structure}. 

\noindent\textbf{Experimental Settings.~} In contrast to the single scale tokenization, the first enhancement exploits a cross-scale embedding layer which composes of four different kernel sizes used in~\cite{wang2021crossformer}. The second variant uses the similar overlapping patch embedding with PVTv2~\cite{wang2021pvtv2}. The third modification further incorporates self-attention that allows for inter-token modeling, as presented in T2T ViT~\cite{yuan2021tokens}. We also compare with the frozen randomly-initialized variant as in MoCov3~\cite{chen2021empirical}. We compare all variants on three tasks, including supervised classification on ImageNet, linear probing with unsupervised pre-training as in~\cite{chen2021empirical}, and semantic segmentation on ADE20K~\cite{zhou2019semantic}. More details including specific designs and experimental settings are provided in Appendix~\ref{app:structure}.

\begin{table*}[htb]
\caption{\textbf{Performance on various vision tasks with different tokenizers.} ``Intra'', ``Local'', and ``Inter'' respectively refer to applying the intra-token, locality, and inter-token refinement strategies. ``Frozen'' refers to the frozen randomly-initialized tokenization in~\cite{chen2021empirical}. }
  \centering
\begin{tabular}{ccccc|cc|cc|c}
\Xhline{1.1pt}
\multirow{2}{*}{Architecture} & \multicolumn{4}{c|}{Tokenization} & \multirow{2}{*}{Params} & \multirow{2}{*}{GFLOPs} & \multicolumn{2}{c|}{Classification} & \multirow{2}{*}{Segmentation} \\ \cline{2-5} \cline{8-9}
                             & Intra     & Local     & Inter    &   Frozen  &              &   &  Linear~\cite{chen2021empirical} & Supervised    &                        \\ \hline
DeiT-S                       & -         & -         & -         &  - &    22.1                    & 4.6                   &  68.1 &  79.8   &       44.0      \\ \hline
\color{mygray}{DeiT-B}                       & \color{mygray}{-}         & \color{mygray}{-}         & \color{mygray}{-}  & \color{mygray}{-}        & \color{mygray}{86.6}                   & \color{mygray}{17.6}                    &  \color{mygray}{69.6}     &   \color{mygray}{81.8}   &      \color{mygray}{45.2}            \\ \hline
DeiT-S                       &            &       &       &  \checkmark  & 22.1                    & 4.6          &  \textbf{69.0}  &  79.4    &      42.9    \\
DeiT-S                       &  \checkmark         &           &       &    & 22.3                    & 4.7                     &   68.4     &   80.5       &    44.3      \\
DeiT-S                       &   \checkmark         &     \checkmark       &        &   & 22.3                    & 5.1                     &   68.5      &  80.9      &  44.6      \\
DeiT-S                       &      \checkmark      &      \checkmark      &       \checkmark  &    & 25.7                    & 6.9                     &  68.7 & \textbf{82.0}    &     45.0     \\ \Xhline{1.1pt}
\end{tabular}
\label{table:structure}
\end{table*}%
 
 \subsection{Takeaways from the Empirical Analysis}
 
To illustrate how the information accessibility between image and tokens changes after tokenization, we follow the practice used in~\cite{hjelm2018learning,vincent2010stacked} and reflect it using reconstruction error. When training completes, we construct an encoder-decoder framework using the pre-trained tokenizer and a random-initialized lightweight decoder. Given the frozen tokenizer, the decoder is then trained to reconstruct the input images in 64$\times$64 using $L_2$ loss for 10 epochs on ImageNet training set. Note that we find the decoder, albeit with a simple three-layer structure, is easy to optimize with good convergence.  In addition to reconstruction, we also measure the token similarity by averaging the cosine similarity between obtained tokens, where lower similarity represents more diversity in the token embedding.

\begin{figure}[htb]
\centering 
\includegraphics[width=\linewidth]{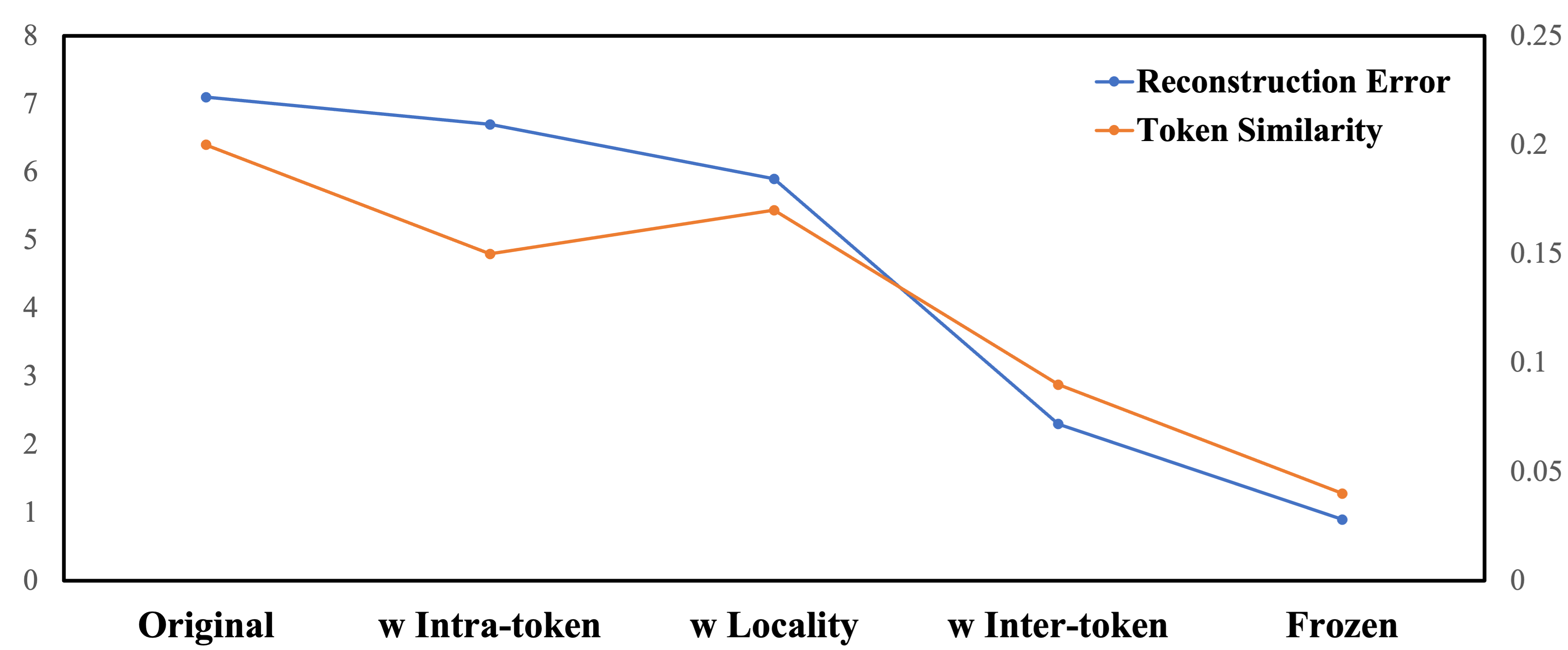}
\caption{\textbf{Reconstruction error and token similarity from different tokenizer designs.} Reconstruction error measures the MSE error from decoding the tokens back to images and reflects their conditional entropy. Token similarity averages the cosine similarity between tokens and reflect the diversity of obtained tokens. Both metrics are evaluated on ImageNet validation set.}
\label{fig:curve}
\end{figure}

From Table~\ref{table:structure} and Figure~\ref{fig:curve}, we perform in-depth analysis and provide several important takeaways for designing better vision tokenizers:
 
\begin{enumerate}
    \item \textbf{Empirical correlations exist between increasing token diversity and better results.} The results in Table~\ref{table:structure} verify that three refinement strategies consistently benefit the transformer's capability. By connecting it with Figure~\ref{fig:curve}, we observe that more diversity and interactions across tokens gradually reduce reconstruction error. 
    \item \textbf{Maximizing conditional entropy doesn't guarantee better performance.} While the randomly-initialized frozen tokenizer or raw pixel encoding maximizes conditional entropy and token diversity, it gives inferior performance on vision tasks. Considering the modality of image, certain properties such as multi-scale are absent with these encodings, which is crucial for vision tasks, especially segmentation.
    \item \textbf{Tokenizers might influence optimization across tasks.} Similar to the findings in~\cite{chen2021empirical}, we also observe better linear performance with frozen tokenizations. However, such trick doesn't apply to supervised classification and segmentation. As has been analyzed in~\cite{xiao2021early}, convolutional stem also provides more stable training process. 
    \item \textbf{The goal of tokenizer is to maintain a trade-off between feature expressiveness and information accessibility.} Unlike the semantically-structured language representations, images consist of sensory pixels and have lower and imbalanced information density. Such characteristics require the tokenizer to perform feature extraction and ``patchifying'' simultaneously, instead of eagerly minimizing the information lost as in language. 
\end{enumerate}
 
Motivated by these findings, we expect our tokenizer to simultaneously maintain the information accessibility and feature quality. Therefore, we further study versatile and unexplored design strategies for vision tokenizers, through normalization and optimization.

\section{\label{sec:norm}Normalization in Vision Tokenizer}
Normalization is vital in network design, which empowers complex architectures and accelerates convergence. Although the importance of normalization has been acknowledged in transformer for both language~\cite{xiong2020layer,shen2020powernorm} and vision~\cite{touvron2021going}, it has rarely been explored inside the tokenizer. As analyzed in Section~\ref{sec:structure} that a good tokenizer is critical for transformer, proper normalization is tailored in this section.

\begin{figure}[htb]
\centering 
\includegraphics[width=\linewidth]{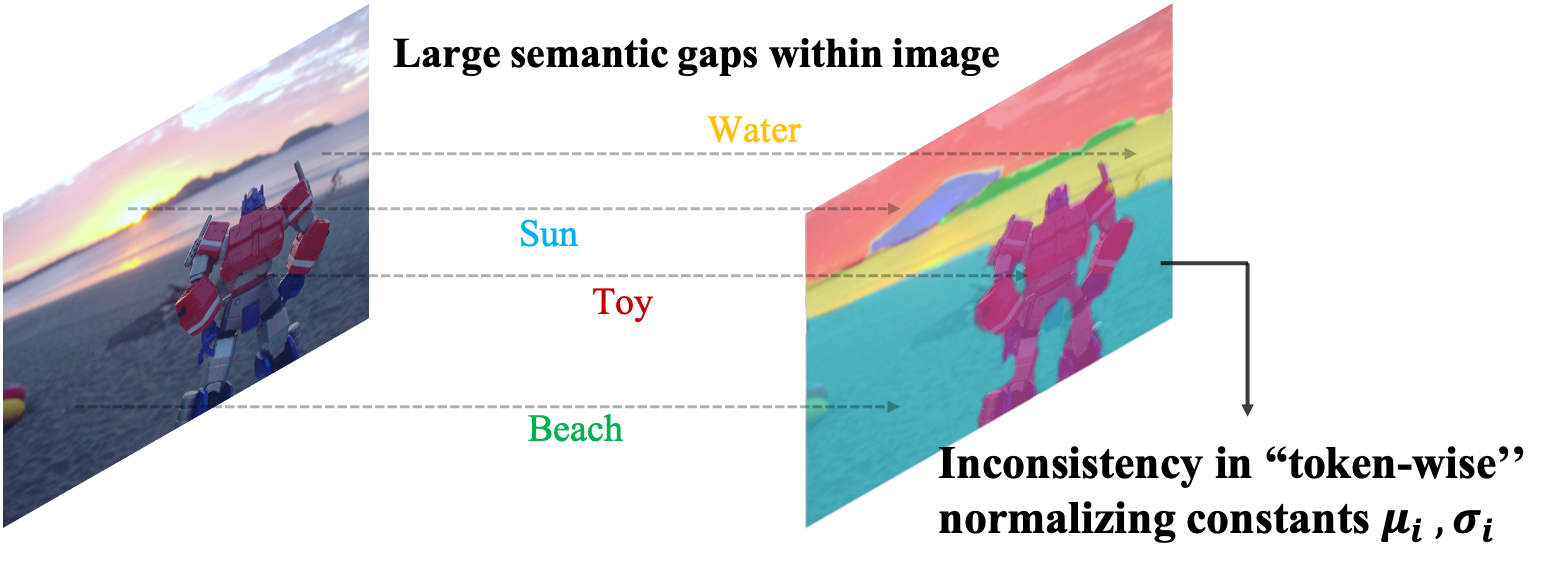}
\caption{\textbf{Example about the large semantic gap across regions and tokens.} This difference leads to distinctive statistics. Using fixed normalizing constants would inevitably ``filter'' the original information and reduces the token diversity.}
\label{fig:image_norm}
\end{figure}

Considering the semantic variations between images, patch tokens tend to encounter semantic variation as shown in Figure~\ref{fig:image_norm}. This large semantic gap across tokens manifests the difficulties in optimizing the tokenizer, which may explain the witnessed training instability of a naive patch embedding~\cite{chen2021empirical,xiao2021early}. However, the addition of batch normalization~(BN) to the patchify stem~\cite{xiao2021early} worsens its accuracy. According to previous analysis~\cite{huang2017arbitrary,park2019semantic}, normalization layers tend to ``wash away'' texture and semantic information. Such ``filtered'' effect inevitably reduces token diversity and possibly cuts off necessary diversity in transformer. Meanwhile, based on our findings in Section~\ref{sec:structure}, naively maximizing token accessibility as in MoCov3~\cite{chen2021empirical} might be suboptimal for different vision tasks. 

\subsection{Modulation Across Tokens~(MoTo)}

\begin{figure}[htb]
\centering 
\includegraphics[width=\linewidth]{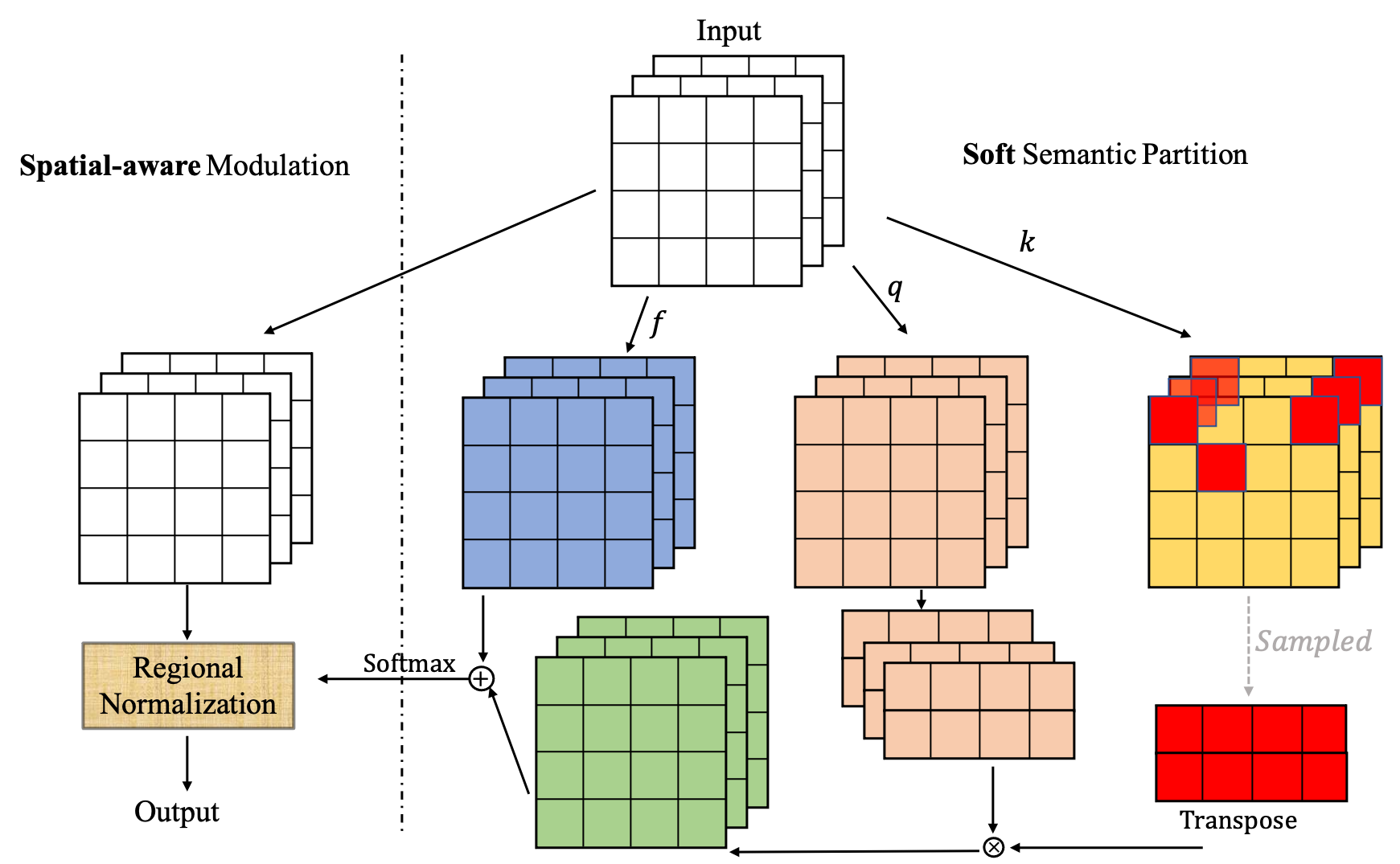}
\caption{\textbf{Forward illustration of MoTo.} The input first proceeds soft semantic partition to obtain the layout. Then spatial-aware modulation is performed based on the soft layout. The soft semantic partition is shown in the right pattern and the spatial-aware modulation is on the left.}
\label{fig:moto}
\end{figure}

Therefore, we propose to normalize the input content in a spatial-aware manner called \textbf{MoTo}, short for \textbf{Mo}dulation across \textbf{To}kens. The core idea is to modulate the diverse semantics in input using regional statistics. This strategy not only formulates feature distributions, but also provides the tokenizer with more plausible semantic content. Unlike the spatial normalization used in conditional image generation~\cite{park2019semantic,miyato2018cgans}, there exists no given semantic layout in our pipeline. As shown in Figure~\ref{fig:moto}, MoTo consists of the soft semantic partition and the spatial-aware modulation.

\subsubsection{Soft Semantic Partition.} Assume each input image consists of $n$ semantic entities, this module predicts the soft semantic layout $\mathbf{L} \in \mathcal{R}^{h \times w \times n}$ from the input $\mathbf{X} \in \mathcal{R}^{h \times w \times c}$. A convolution layer $f$ with $n$ output channels first extracts the semantic feature $f(\mathbf{X})\in \mathcal{R}^{h \times w \times n}$. Similarly, two feature extractors $k$ and $q$ obtain $k(\mathbf{X})$ and $q(\mathbf{X})$. 
Then $n$ feature points $k_n(\mathbf{X})$ are uniformly sampled from $k(\mathbf{X})$ following the practice in~\cite{caron2018deep} to compute the feature correlation. The semantic activation map $\mathbf{Z} \in \mathcal{R}^{h \times w \times n}$ is calculated using matrix multiplication and accumulated with semantic feature

\begin{equation}
\mathbf{Z} = \mathbf{u} \cdot k_n(\mathbf{X})^T q(\mathbf{X}) + f(\mathbf{X}),
\end{equation}
\noindent where $\mathbf{u} \in \mathcal{R}^{1 \times 1 \times n}$ denotes a randomly-initialized learnable dictionary that integrates the correlation with features. 

Then $\mathbf{Z}$ is normalized using softmax to generate the soft layout at $k$-th semantic entity 

\begin{equation}
\mathbf{L}_k = \frac{\exp (\tau \mathbf{Z}_k)}{\sum_{i=1}^{n} \exp (\tau \mathbf{Z}_i)},
\label{eq:softmax}
\end{equation}
where $k \in [0,n-1]$ and $\tau$ is the temperature coefficient set to 0.1. Each $\mathbf{L}_k \in \mathcal{R}^{h \times w}$ indicates the probability of the spatial pixels belonging to entity $k$.

\begin{table*}[htb]
\caption{\textbf{Performance of image recognition on ImageNet validation set with MoTo}. The column of ``Params'' denotes the number of parameter, ``Acc'' reports Top-1 accuracy. All experiments and GFLOPs computations use the input size of 224 $\times$ 224.}
\label{table:sota}
\centering
\resizebox{0.6\linewidth}{!}{
\begin{tabular}{c|c|cc|cc}
\Xhline{1.1pt}
\begin{tabular}[c]{@{}c@{}}Transformer \\ Architecture\end{tabular}                & Model                 & Params & GFLOPs & Accuracy & $\Delta$ \\ \Xhline{1pt}
\multirow{4}{*}{ViT~\cite{dosovitskiy2021an,touvron2020training} } & DeiT-S                & 22.1M           &          4.6               & 79.8   &    -    \\
                                  & \textbf{w MoTo}                &      22.5M      &                        4.8                            &     \textbf{81.6}  &  +1.8  \\ \cline{2-6} 
                                  & DeiT-B         &      86.6M          &                          17.6                                       &    81.8    &   -        \\
                                  & \textbf{w MoTo}          &     86.9M       &                      17.9                                     &       \textbf{82.9}    & +1.1   \\ \hline
\multirow{2}{*}{T2T-ViT~\cite{yuan2021tokens}}   & T2T-ViT-14            & 21.5M          &    5.2                                                     & 81.5    &  -     \\ 
                                  & \textbf{w MoTo}     &      21.8M       &             5.4                                                    &       \textbf{82.3}  & +0.8   \\ \hline
\multirow{6}{*}{PVT~\cite{wang2021pyramid,wang2021pvtv2}} & PVT-Small               & 24.5M           &        3.8                                                         & 79.8   &   -     \\
                                  & \textbf{w MoTo}                 &   24.7M        &          4.0                                                       & \textbf{81.0}   &   +1.2     \\ \cline{2-6} 
                                  & PVT-Medium        &       44.2M        &                   6.7                                             &   81.2   &   -      \\
                                  & \textbf{w MoTo}         &     44.5M          &          6.9                                                     &       \textbf{82.1}   &   +0.9  \\ \cline{2-6}
                                  & PVTv2-B2        &       25.4M        &                   4.0                                             &   82.0   &   -      \\
                                  & \textbf{w MoTo}         &    25.6M          &          4.2                                                     &       \textbf{82.7}  &   +0.7     \\ \hline                                 
\multirow{4}{*}{Swin~\cite{liu2021swin}} & Swin-T                & 28.3M           &        4.5                                                         & 81.2     &   -       \\
                                  & \textbf{w MoTo}                 &   28.6M        &          4.7                                                       & \textbf{82.2}     &   +1.0      \\ \cline{2-6} 
                                  & Swin-S        &       49.6M        &                   8.7                                             &   83.0    &   -     \\
                                  & \textbf{w MoTo}         &     49.9M          &          8.9                                                     &       \textbf{83.7}  &  +0.7  \\ \Xhline{1.1pt}
\end{tabular}}%
\end{table*}%

\subsubsection{Spatial-aware Modulation.} 
As the soft layout provides distributions of potential semantic entities, this component modulates the input with regional normalization. In contrast to instance normalization~\cite{ulyanov2016instance,huang2017arbitrary}, it models the spatial correlations and treats semantic entities as ``instances''. Formulating similar semantics with shared means and variances, our method modulates the interactions in input better, as 

\begin{equation}
\mathbf{\text{MoTo}(X)} = \sum_{i=1}^{n} (\frac{\mathbf{X}-\mu(\mathbf{X} \odot \mathbf{L}_i )}{\sigma(\mathbf{X} \odot \mathbf{L}_i )+\epsilon} \times \mathbf{\beta}_{i} + \mathbf{\alpha}_{i}) \odot \mathbf{L}_i,
\label{eq:norm}
\end{equation}
\noindent where $\mu(\cdot)$ and $\sigma(\cdot)$ respectively denote computing the mean and standard deviation from the selected instance. $\mathbf{\beta}_{i}$ and $\mathbf{\alpha}_{i}$ in Eq.~\ref{eq:norm} are the learnable parameters for affine transformation in normalization layers. We use a default $n = 8$ in most experiments.

\subsection{\label{sec:motoexp}Experimental Analysis with MoTo}

\subsubsection{Improvements on various architectures.} To further validate that MoTo is versatile with different tokenizers and transformer architectures, we integrate the proposed module to several state-of-the-art methods, including DeiT~\cite{touvron2020training}, T2T-ViT~\cite{yuan2021tokens}, PVT~\cite{wang2021pyramid}, and Swin Transformer~\cite{liu2021swin}. Since these frameworks contain distinctive training pipelines, we utilize their publicly-available scripts~\cite{rw2019timm,code2021deit,code2021swin,code2021T2T,code2021PVT} and keep the training parameters consistent with the provided ones. When applied with our module, the variants maintain the consistent configurations including augmentation and optimizer as the baselines. We train all models on ImageNet~\cite{deng2009imagenet} training set using 8 Tesla V100 GPUs for 300 epochs, except for T2T-ViT which originally scheduled 310 epochs. From Table~\ref{table:sota}, we see that the consistent improvements are made across different transformer architectures. For the wide-adopted baseline DeiT~\cite{touvron2020training}, our strategy improves its naive tokenization process.The improvements can be observed on both PVT~\cite{wang2021pyramid} and PVTv2~\cite{wang2021pvtv2}, where PVTv2 introduces overlapping patch embedding.  While T2T-ViT exploits a transformer in tokenizer to perform re-structurization, MoTo is still effective. 

\vspace{0.05in}
\noindent\textbf{Complexity Analysis.~~} MoTo incorporates global context modeling to vision tokenizer using normalization. The channel, height, weight of the feature maps and number of semantic entities are respectively denoted as $C,H,W,N$. Implemented with convolution layers and sampling, soft semantic partition and spatial-aware modulation cost a time complexity of $\mathcal{O}(NCHW)$. Comparing to the $\mathcal{O}(C(HW)^2)$ complexity of self-attention module, MoTo provides a computational-friendly choice to perform inter-token modulation. As shown in Table~\ref{table:speed}, the actual computation of injecting pixel-wise regional information with self-attention scales quadratically, which is unfeasible for processing high-resolution images. 

\begin{table}[htb]
\caption{\textbf{Inference wall time~(ms) with different input scales.} Features are fed into the same GPU with a batch size of 1 and channel number of $16$. OOM denotes out-of-memory. Note that the definition of self-attention here pixel refers to the pixel-wise self-attention in tokenizer.}
\label{table:speed}
  \centering
 \resizebox{\linewidth}{!}{
\begin{tabular}{c|cccc}
\Xhline{1.1pt}
Component       & $224 \times 224$ & $384 \times 384$ & $512 \times 512$ & $1024 \times 1024$ \\ \hline
Self-attention &  4.12   &  23.58   & OOM & OOM  \\ 
MoTo           &  0.52   &  1.67   &    4.23  & 18.42 \\ \Xhline{1.1pt}
\end{tabular}}%
\end{table}%

\subsubsection{Visualization of semantic layout.} 
\begin{figure}[htb]
\centering 
\includegraphics[width=\linewidth]{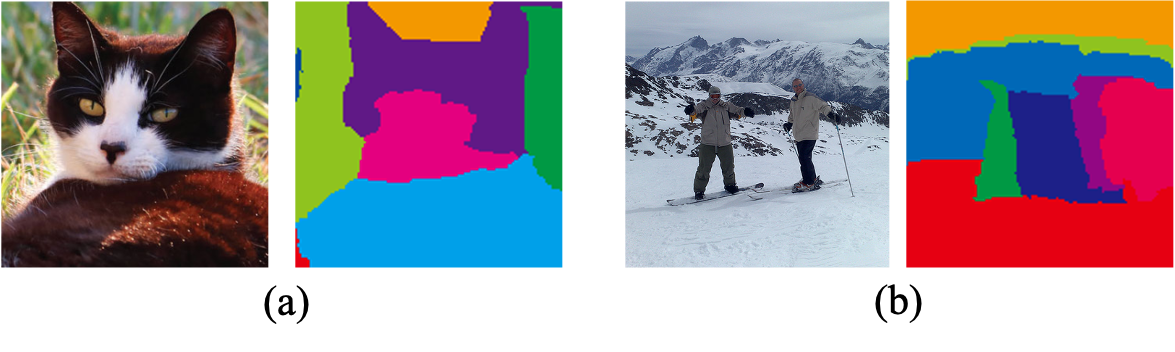}
\caption{\textbf{Visualizations of the self-learned semantic layout.} Both image (a) and (b) are testing images. Each color shown in the layout denotes a semantic entity in the soft layout $\mathbf{L}$.}
\label{fig:layout}
\end{figure}

To understand the effectiveness of the semantic partition, we visualize the obtained semantic entities using a $n$-color palette. Given its soft semantic vector $\mathbf{L}_{i,j} \in \mathcal{R}^{1 \times 1 \times k}$, we colorize each location ($i,j$) using the index with the highest probability. As shown in Figure~\ref{fig:layout}, the semantic entities highlight the difference between foreground, background, and instances. Notably, the entities represent some details on cat face in Figure~\ref{fig:layout}.(a), which switch into instances such as human and mountains in Figure~\ref{fig:layout}.(b). The results demonstrate the effectiveness of our module in capturing semantic variations in images. Normalizing these feature points with semantic grouping improves the feature quality of tokens. 

\begin{table}[htb]
\caption{\textbf{Ablation study about entity numbers and parition strategy}. The baseline architecture uses DeiT-S. Each row represents a model trained with different number of semantic entities or partition strategy. The gray rows refer to models with hard partition.}
\label{table:partition}
\centering
\resizebox{0.9\linewidth}{!}{
\begin{tabular}{c|c|cc}
\Xhline{1.1pt}
\multirow{2}{*}{Semantic Entities} & \multirow{2}{*}{Partition Strategy} & \multicolumn{2}{c}{Top-1 Accuracy} \\
                                   &                                     & Val           & $\Delta$           \\ \hline
-                 & -         &     79.8      &  -   \\ \hline
$n$ = 2                 & Soft         &   80.8      &  +1.0     \\ 
$n$ = 4                 & Soft         &    81.1     &      +1.3 \\ 
$n$ = 8                 & Soft         &    81.6     &  +1.8     \\ 
$n$ = 16                 & Soft         &    81.7     &  +1.9      \\ \hline
\color{mygray}{$n$ = 16}                  &  \color{mygray}{Hard}         &   \color{mygray}{80.5}    &    \color{mygray}{+0.5}     \\ 
\color{mygray}{$n$ = 32}                  &  \color{mygray}{Hard}         &     \color{mygray}{80.8}  &     \color{mygray}{+0.8}    \\ \Xhline{1.1pt}
\end{tabular}}%
\end{table}%

\subsubsection{Ablations about MoTo.}

\vspace{0.05in}
\noindent\textbf{Number of semantic entities.~~} The hyper-parameter $n$ number of the semantic entities determines our semantic partition process. We gradually increase $n$ to study the influence of varying quantities of semantic entities. When the number is small, Table~\ref{table:partition} shows a trend of improved accuracy with more entities. However, growing grouping numbers bring relatively marginal improvements when $n > 8$. As the quantity controls the fineness of semantic modeling, increasing entities contribute to stronger context modeling while inevitably bring redundancy and difficulties in optimization, unable to guarantee better results. We thus choose $n = 8$ for most experiments. considering the trade-off between effectiveness and complexity. 

\vspace{0.05in}
\noindent\textbf{Soft and hard partition strategy.~~} To understand our partition strategy, we also compare the difference between soft and hard partition. In contrast to the soft probability layout in Eq.~\ref{eq:softmax}, we implement $\argmax$ on the activation maps spatially and further obtain a class map to perform hard partition, which resembles the visualization process in Figure~\ref{fig:layout}. As shown in Table~\ref{table:partition}, hard partition performs much worse than soft partition even with more semantic entities, exhibiting the superiority of soft partition.

\vspace{0.05in}
\noindent\textbf{Absorbing MoTo into transformer blocks.~~} While we are mainly discussing replacing the original patch embedding layer with better tokenization, it remains unexplored to absorb such tokenization into the transformer blocks. We further conduct a pilot study to ensemble MoTo at the end of each transformer layer in Table~\ref{table:absorb}. Despite with clear improvements with MoTo as tokenization and increased complexity, fusion into transformer blocks gives limited elevations.  
We can also see the importance of tokenizer-level normalization, apart from its follow-up transformer blocks. The results not only suggest the prominence of vision tokenization, but also elaborate more towards the information flow of transformer. 

\begin{table}[htb]
\caption{\textbf{Ablation study on absorbing MoTo into transformer blocks}. The baseline architecture uses DeiT-S. We adopt a MoTo layer with $8$ semantic entities and ensemble it into transformer blocks. The placement denotes the layer number of transformer block that adopts MoTo.}
\label{table:absorb}
\centering
\resizebox{0.9\linewidth}{!}{
\begin{tabular}{cccc|cc}
\Xhline{1.1pt}
& \multicolumn{3}{c}{Placement} & \multicolumn{2}{c}{Top-1 Accuracy} \\ \cline{2-6}
    Tokenizer & 1-4    & 4-8 & 8-12  & Val           & $\Delta$           \\ \hline
-   &   -   &   -      & -         &     79.8     &  -   \\ \hline
$\checkmark$   &      &         &          &   81.6      &  +1.8     \\ 
$\checkmark$   &  $\checkmark$   &        &          &    81.8     &      +2.0 \\ 
$\checkmark$  &  $\checkmark$  &  $\checkmark$    &      &    81.9     &  +2.1     \\ 
$\checkmark$ & $\checkmark$  & $\checkmark$    & $\checkmark$     &    81.4     &  +1.6      \\ \Xhline{1.1pt}
\end{tabular}}%
\end{table}%

\subsubsection{Discussions on normalization.} Image synthesis~\cite{park2019semantic,wang2020attentive,huang2017arbitrary} and domain adaptation~\cite{pan2018two,li2016revisiting,wu2018dcan}, where large domain gap and variations exist, recently find normalizations benefit the networks in encoding rich content. Motivated by these findings, our method investigates normalization inside vision tokenizer and supplements tokenization process with capability of context modeling. The clear boost over multiple structures brought by this lightweight design further demonstrates the importance of a good tokenizer. 

In contrast to normalization layers inside transformer blocks, our work investigates the normalization inside vision tokenizers. We emphasize that these two directions are inherently different and both crucial, while the modulation in tokenizers has been rarely explored. 

\vspace{0.05in}
\noindent\textbf{Normalization in Transformer.~~} Previous works have demonstrated the importance of normalizations in transformers~\cite{vaswani2017attention,xiong2020layer}, where Layer Normalization~\cite{ba2016layer} plays a key role in their success. With further developments in normalization such as PowerNorm~\cite{shen2020powernorm}, transformers on language modeling have received a performance boost. As the paradigm shift becomes success in computer vision, analysis and attempts on normalizations in vision transformers have been made in~\cite{yao2021leveraging,touvron2021going}. 

\vspace{0.05in}
\noindent\textbf{Normalization in Tokenizer.~~} Unlike the normalizations conducted inside transformers which are capable of maintaining the attention magnitudes, normalizations in tokenizers perform a different role of token feature extraction. In language pipelines, normalizations in tokenizers refer to the operations that make a raw string cleaner, including traditional Unicode normalization and BERT normalizer~\cite{devlin2018bert}. Comparing to highly structured languages, images contain more complex semantics and diversified pixel variations. Thus tokenization on images is more challenging and might require additional regularization. However, normalizations in vision tokenizers have been rarely explored. Our proposed MoTo targets at fixing this omitted ingredient in vision tokenizer. 

Based on the different roles between normalization in tokenizers and transformers, their design strategies should be made different accordingly. Considering the semantic variations across images and tokens, MoTo normalizes the input content in a spatial-aware manner that modulates the regional variations while not ``wash away'' the diverse texture and semantic content of inputs. Based on the analysis in our main paper, MoTo is capable of maintaining the information accessibility and diversity between input images and extracted tokens. With minimal computational costs through modulations, MoTo also incorporates global information into the tokenization process.

\vspace{0.05in}
\noindent\textbf{Comparison with different standard normalizations.~~} 
To better illustrate the effectiveness of MoTo, we present more ablative analysis by comparing MoTo to other choices of normalizations in tokenizers. We use DeiT-S architecture as our baseline and follow the experiment settings in Section~\ref{sec:motoexp}. We choose Layer Normalization~(LN), Batch Normalization~(BN) and instance normalization~(IN) as the counterparts of MoTo.

\begin{table}[htb]
\caption{\textbf{Comparison of different normalizations in tokenizer.} The baseline architecture uses DeiT-S. Each row denotes result trained with different normalization strategy. Top-1 Accuracy denotes the validation accuracy on ImageNet.}
\centering
\resizebox{\linewidth}{!}{
\begin{tabular}{c|c|c}
\Xhline{1.1pt}
Model  & Normalization in Tokenizer & Top-1 Accuracy \\ \hline
DeiT-S & -                 & 79.8           \\ \hline
DeiT-S & Layer Norm        & 79.6           \\
DeiT-S & Batch Norm        & 79.5           \\
DeiT-S & Instance Norm     & 80.1           \\ \hline
DeiT-S & \textbf{MoTo}              & \textbf{81.6}           \\ \Xhline{1.1pt}
\end{tabular}}%
\label{table:norm}
\end{table}%

As shown in Table~\ref{table:norm}, both BN and LN slightly harm the model's performance. As analyzed, such normalization methods bring ``filtered'' effect on tokens' features. In comparison with IN, MoTo performs soft semantic partition and incorporates spatial-aware property into the instance-wise modulation. The results further indicate that different from normalization in transformer blocks, normalization in tokenizers need to take both ``intra-token'' and ``inter-token'' modeling into considerations.

\section{Objectives for Vision Tokenizer}
From the information trade-off perspective, both structural modifications in Section~\ref{sec:structure} and MoTo in Section~\ref{sec:norm} can be considered as refinements that assign the tokenizer with stronger capability in preserving the conditional entropy between image and token representations. In addition to network-level analysis, we explore optimization-level refinement for vision tokenizers in this section.

\subsection{The ``greedy'' training paradigm.} Recent studies~\cite{wang2021revisiting,belilovsky2019greedy} have observed a ``greedy'' trend in the models trained with supervised signals. The network is optimized to encode task-relevant features under the guidance of a loss function, e.g. cross-entropy loss for classification. As the latent features become more discriminative, the conditional entropy between inputs and layer-wise features gradually decreases during training according to the estimation in~\cite{wang2021revisiting}. This ``greedy'' characteristic in locally-supervised learning pipeline often collapses task-relevant feature in earlier layers, which further leads to inferior performance. Although not pushed by additional supervision, the tokenizer inevitably reduces information accessibility between the input and tokens during training. Considering the difficulty of vision tokenization and witnessed instability~\cite{chen2021empirical}, this ``greedy'' phenomenon possibly exists in vision tokenizers, washing out some valuable information. With similar observations in language understanding~\cite{kong2019mutual,voita-etal-2019-bottom}, a potential solution is to modify the objectives to regularize the tokenizer from being ``short-sighted''. 

The objective of Masked language modeling~(MLM) in BERT~\cite{devlin2018bert} has been demonstrated with its property in guiding the evolution of representations of individual tokens proceed in two stages, including context modeling and token reconstruction~\cite{voita-etal-2019-bottom}. InfoWord~\cite{kong2019mutual} further analyzes how MLM differs from InfoNCE~\cite{gutmann2012noise} and incorporates negative sampling in measuring more concise mutual information, hereby learning better language representations. 

Therefore, we attempt to design a less ``greedy'' training regime, where the \textit{head}, vision tokenizer is able to preserve more context from inputs that can potentially leverage by later \textit{body}, transformer blocks.

\subsection{TokenProp Objective}

 While performing different roles in the information flow of transformers, the vision tokenizer and transformer blocks have been treated equally during training. The core idea of TokenProp is to provide optimization objectives for vision tokenizer, which lead to better token representations by maintaining the trade-off between feature fineness and information accessibility.

Following the notations in Section~\ref{sec:structure} where $\mathbf{A}$ and $\mathbf{B}$ refer to the input image and tokens, we show that the conditional entropy between them could be maximized by optimizing $\mathbb{R}(\mathbf{A}|\mathbf{B})$

\begin{equation}
\begin{aligned}
    \argmax{I(\mathbf{A};\mathbf{B})} & = \argmax{-\mathbb{R}(\mathbf{A}|\mathbf{B})} \\
                    & = \argmax{\mathbb{E}_{q(\mathbf{A},\mathbf{B})}[\log q(\mathbf{A}|\mathbf{B})]}
    \end{aligned}
\label{eq:mutual-argmax}
\end{equation}
where we estimate the parameters of $q(\mathbf{A}|\mathbf{B})$ using a decoder network $G_{\theta}$. Ideally, zero information loss through the tokenization process retains all information including the useful one. 

However, solely optimizing the conditional entropy collapses into another extreme case, where the tokenizer merely performs spatial partition and doesn't capture any task-specific feature. Therefore, we utilize a locally-supervised paradigm that jointly optimizes the target loss and the conditional entropy during tokenization. The surrogate optimization objective could be defined as

\begin{equation}
    \underset{\phi,\omega,\theta}{\text{minimize}} \mathcal{L}(F_{\omega}(F_{\phi}(\mathbf{x}));\textbf{y}) + \lambda \mathcal{L}_{rec} (G_{\theta}(F_{\phi}(\mathbf{x})); \textbf{x})
\label{eq:loss-tokenprop}
\end{equation}
where $\mathcal{L}_{rec}$ and $\mathcal{L}$ represent the reconstruction loss and the standard task loss, e.g.\ cross entropy loss for classification in our case. $F_{\phi}$ and $F_{\omega}$ denote the tokenizer and the follow-up transformer architecture. $\textbf{x}$, $\textbf{y}$, $\phi$, $\omega$, $\theta$ refer to the input, label, and parameters of respective modules.

\subsection{Experiments with TokenProp}

\vspace{0.05in}
\noindent\textbf{Implementation Details.~~} We adopt a lightweight decoder structure to reconstruct the input from tokens. We find that a simple three-layer decoder works well in most cases. The decoder first transforms the split tokens into connected spatial feature maps, which are further refined using convolutional layers and upsampled to $64 \times 64$. This implementation introduces negligible computational overhead to the transformer pipeline. The default hyper-parameter $\lambda$, which combines standard loss and reconstruction loss, is set 0.001 in our experiments. All the experiments are trained for 300 epochs on 8 GPUs if not additionally specified.

\vspace{0.05in}
\noindent\textbf{Choices of $\mathcal{L}_{rec}$.~} The reconstruction loss $\mathcal{L}_{rec}$ evaluates the conditional entropy between image and token representations. As there exists multiple choices to compute the pixel-wise distance, we perform a comparison in Table~\ref{table:loss}. Apart from the commonly used $L_1$ and $L_2$, we also adopt the perceptual loss~\cite{johnson2016perceptual} and contextual loss~\cite{mechrez2018contextual}. Interestingly, the $L_2$ distance shows a competitive result over other alternatives. While perceptual loss excels at capturing style information and better visual quality~\cite{johnson2016perceptual,zhu2017unpaired}, it doesn't perform well as expected. Contextual loss, which is effective at maintaining pixel statistics, boosts slightly more than $L_2$. Note that both perceptual loss and contextual loss exploit a pre-trained VGG-19, which might bring information leakage into training. Considering the complexity of contextual loss,  we use $L_2$ distance in our experiments. Meanwhile, the results indicate that better reconstructed quality doesn't necessarily mean higher accuracy.

\begin{table}[htb]
\caption{\textbf{Ablations about $\mathcal{L}_{rec}$.} The baseline architecture uses DeiT-S. Each row denotes result trained with different $\mathcal{L}_{rec}$.}
\label{table:loss}
\centering
\resizebox{0.9\linewidth}{!}{
\begin{tabular}{c|c|cc}
\Xhline{1.1pt}
\multirow{2}{*}{Model} & \multirow{2}{*}{Loss Type} & \multicolumn{2}{c}{Top-1 Accuracy} \\ \cline{3-4} 
             &              & Val           & $\Delta$           \\ \hline
Baseline    &     -                  &      79.8         &      -         \\ \hline
-        &   $L_1$                &         80.2      &     +0.4      \\
-  &            $L_2$               &       80.7        &   +0.9            \\
-  &        Perceptual Loss~\cite{johnson2016perceptual}               &        80.4       &        +0.6       \\
-  &        Contextual Loss~\cite{mechrez2018contextual}              &        80.8       &         +1.0      \\ \Xhline{1.1pt}
\end{tabular}}%
\end{table}%

\vspace{0.05in}
\noindent\textbf{Reconstruction loss weight of $\lambda$.~} As the target features for reconstruction and classification are completely different, the reconstruction loss weight $\lambda$ servers as another crucial hyper-parameters. If the model focuses too much on reconstruction, the final classification performance will inevitably be affected. We perform a study on how a different $\lambda$ influences our model. We adopt different loss weights of 0.001, 0.01, 0.1 and 1.0 and report the respective performance of these four variants. The training recipes are kept the same as the initial one. From Table~\ref{table:weight}, we can find that our method produces similar improvements when given different reconstruction weights at a reasonable range. If the weight is set too large, instability will be observed in training. The results also demonstrate the robustness of TokenProp and indicates the potential linkage between generative and discriminative signals.

\begin{table}[htb]
\caption{\textbf{Ablations about the reconstruction loss weights $\lambda$.} The baseline architecture uses DeiT-S. Each row denotes result trained with different $\lambda$. NaN means the model faces NaN in training loss.}
\label{table:weight}
\centering
\resizebox{0.7\linewidth}{!}{
\begin{tabular}{c|c|cc}
\Xhline{1.1pt}
\multirow{2}{*}{Model} & \multirow{2}{*}{Loss Weight $\lambda$} & \multicolumn{2}{c}{Top-1 Accuracy} \\ \cline{3-4} 
             &              & Val           & $\Delta$           \\ \hline
Baseline    &     -                  &      79.8         &      -         \\ \hline
-        &   0.001                &         80.7      &     +0.9      \\
-  &            0.01               &       80.4        &   +0.6            \\
-  &        0.1               &        80.6       &        +0.8      \\
-  &        1.0              &        NaN       &         -      \\ \Xhline{1.1pt}
\end{tabular}}%
\end{table}%

\vspace{0.05in}
\noindent\textbf{Decoder Structure.~} We also study whether a complex decoder is favoured in our setting. By enlarging the convolutional channel for $n$ times, we denote the decoder variant as $\times n$. We also compute reconstruction loss using outputs with higher resolutions, which are obtained by stacking more upsampling layers to the decoder. The training overhead is measured on the same 8 GPUs by comparing the final 300 epochs' training time with the baseline trained without TokenProp. While more complex decoders introduce extra overhead in Table~\ref{table:decoder}, we see no clear improvement with larger decoders. Similar with sophisticated designs, reconstructions with higher resolutions tend to preserve more details. However, a performance drop is observed when we use the output size of 256 $\times$ 256, which emphasizes the importance of a proper objective. These comparisons, as well as findings from Table~\ref{table:loss}, also indicate that a simple decoder is sufficient for learning our objective, which is not designated for generating high quality outputs. 

\begin{table}[htb]
\caption{\textbf{Analysis about the decoder structures.} The baseline architecture uses DeiT-S. The sign of $\downarrow$ denotes the accuracy is lower than the baseline trained without TokenProp.}
\label{table:decoder}
\centering
\resizebox{\linewidth}{!}{
\begin{tabular}{cc|c|c}
\Xhline{1.1pt}
Decoder Channel      & Output Scale                 & Accuracy              & Training Overhead    \\ \hline
$\times 1$                   & 64 $\times$ 64                   & 80.7                  &        4.1\%              \\ \hline
$\times 1$                   &  128 $\times$ 128                     &        80.4               &       7.8\%               \\
$\times 1$                   &  256 $\times$ 256                     &        77.3 ($\downarrow$)               &       15.1\%               \\
$\times 2$                   &  64 $\times$ 64                     &        80.4               &       6.9\%              \\
$\times 4$                   &  64 $\times$ 64                     &        80.5               &       9.1\%              \\ 
\Xhline{1.1pt}
\end{tabular}}%
\end{table}%

\begin{table*}[htb]
\caption{\textbf{Performance of image recognition on ImageNet validation set with TokenProp}. The column of ``$\Delta$'' reports the improvements from the models trained with TokenProp.}
\label{table:sota_loss}
\centering
\resizebox{0.6\linewidth}{!}{
\begin{tabular}{c|c|cc|c}
\Xhline{1.1pt}
\begin{tabular}[c]{@{}c@{}}Transformer \\ Architecture\end{tabular}                & Model                 & Accuracy & $\Delta$ & Training Overhead \\ \Xhline{1pt}
\multirow{2}{*}{ViT~\cite{dosovitskiy2021an,touvron2020training} } & DeiT-S                & \textbf{80.7}           &          +0.9               & 4.1\%           \\
                                  & DeiT-B         &      \textbf{82.5}          &                          +0.7                                       &    3.5\%            \\ \hline
\multirow{2}{*}{T2T-ViT~\cite{yuan2021tokens}}   & T2T-ViT-14            & \textbf{82.2}          &    +0.7                                                     &  4.9\%           \\ 
                                                  & T2T-ViT-19            & \textbf{82.8}          &    +0.9                                                     &  4.4\%           \\  \hline
\multirow{2}{*}{PVT~\cite{wang2021pyramid,wang2021pvtv2}} & PVT-Small               & \textbf{80.9}           &        +1.1                                                         & 5.1\%  \\
                                  & PVT-Medium        &       \textbf{81.8}        &        +0.6                                                      &   4.6\%        \\ \hline    
\multirow{1}{*}{Swin~\cite{liu2021swin}} & Swin-T                & \textbf{82.3}           &       +1.1                                                         &  6.7\%           \\
\Xhline{1.1pt}
\end{tabular}}%
\end{table*}%

\subsubsection{Improvements on Various Architectures.} In order to validate the generalization of TokenProp, we apply it to various transformer architectures. We maintain the original training recipes of these state-of-the-arts as mentioned in Section~\ref{sec:motoexp}. Except for the training-only decoder, our method incurs no extra parameter or computation to the original framework. Therefore, these models receive negligible overhead to compute our TokenProp objective with a light-weight decoder during training. As shown in Table~\ref{table:sota_loss}, different models in four families of transformer architectures receive a steady improvement on accuracy, with roughly 5\% training overhead. From another perspective, TokenProp serves as a regularization term that provides the encoded features from tokenizers with more diversity. 

\subsubsection{Compatibility with Optimizers.} Instability is another major curse for vision transformers. AdamW~\cite{loshchilov2018fixing} optimizer has been dominant in mitigating the training difficulty of transformers. Recent findings ascribe transformers' vulnerability to their tokenizers, and try to provide a fix, such as the convolutional stem in~\cite{xiao2021early} and the random patch projection~\cite{chen2021empirical} for self-supervised learning. While these modifications make transformers less sensitive towards training recipes, the AdamW counterparts still exhibit considerable superiority over the ones trained with simpler optimizer such as SGD~\cite{touvron2020training,xiao2021early}. To testify the influence of TokenProp on optimizations, we propose to validate our method using SGD in Table~\ref{table:optim}. We also re-implement a DeiT-S variant with the convoltuional stem in~\cite{xiao2021early}, as well as its combination with TokenProp. Since the frozen randomly-initialized embedding~\cite{chen2021empirical} has been adopted with better stability, we also include such variant with replaced optimizers. We tune the optimal learning rate and weight decay for each variant through multiple runs, while other training recipes such as augmentation are kept consistent. As shown in Table~\ref{table:optim}, TokenProp enables the adaptability of the model towards SGD, which significantly reduces the performance drop. Notably, by incorporating TokenProp with the convolutional stem in~\cite{xiao2021early}, the variant only loses $0.2\%$ validation accuracy. These results further suggest that an optimized objective provides more reliable supervisory signals, which leads to both better performance and stability. Comparing to the additional memory costs from Adam-based methods, TokenProp only incurs limited computation and memory overhead during training. 

\vspace{0.05in}
\noindent\textbf{Comparison with frozen embedding~\cite{chen2021empirical}.~} Although the practice in~\cite{chen2021empirical} observes with better stability in self-supervised learning, we observe inferior performance on supervised classification and downstream tasks in Section~\ref{sec:structure}, as well as heavy reliance on optimizers. We believe there exists a connection between the frozen embeddings and our TokenProp objective, as both methods target to maintain the stability in the tokenization process. Nevertheless, feature fineness is neglected in such raw projection, which accounts for its poor performance on downstream tasks. In contrast, TokenProp serves as a surrogate objective that balances optimization stability and representation quality.

\begin{table}[htb]
\caption{\textbf{Analysis about the compatibility with optimizers.} The sign of $\downarrow$ shows how much the accuracy is lower than the baseline trained using AdamW. We re-implement DeiT with~\cite{xiao2021early} and~\cite{chen2021empirical}, as denoted by DeiT$_{C}$* and w Frozen. We highlight top-2 results in bold font.}
\label{table:optim}
\centering
\resizebox{0.9\linewidth}{!}{
\begin{tabular}{c|c|cc}
\Xhline{1.1pt}
\multirow{2}{*}{Model Variants} & \multirow{2}{*}{Optimizer} & \multicolumn{2}{c}{Top-1 Accuracy} \\ \cline{3-4} 
                                &                            & Val          & $\Delta$          \\ \hline
\color{mygray}{DeiT-S}                          & \color{mygray}{AdamW}                      &       \color{mygray}{79.8}       &      \color{mygray}{-}          \\
DeiT-S                          & SGD                        &       76.7       &   $\downarrow$3.1             \\ \hline
DeiT-S w Frozen~\cite{chen2021empirical}                          & AdamW                      &       79.4       &     -          \\
DeiT-S w Frozen~\cite{chen2021empirical}                          & SGD                      &       76.0       &      $\downarrow$3.4          \\ \hline
DeiT$_{C}$-S*~\cite{xiao2021early}                          & SGD                        &    78.2          &      $\downarrow$1.6          \\ \hline
\textbf{DeiT-S w TokenProp}               & SGD                        &       78.9       &       \textbf{$\downarrow$0.9}        \\
\textbf{DeiT$_{C}$-S* w TokenProp}                 & SGD                        &    79.6          &      \textbf{$\downarrow$0.2}          \\ \Xhline{1.1pt}
\end{tabular}}%
\end{table}%

\section{Discussions.} 
To summarize, we propose two plug-and-play designs in vision tokenizers: Modulation across Tokens~(MoTo) that incorporates inter-token modeling capability through normalization, and a surrogate optimization objective TokenProp. In this section, we provide experimental discussions to further demonstrate the effectiveness of our method. 

\subsection{Ensembles of MoTo and TokenProp.}
Since we've demonstrated that both MoTo and TokenProp are versatile with different transformer architectures, we also validate whether they benefit from each other as well. In Table~\ref{table:combine}, we ensemble both strategies into different frameworks and follow the aforementioned training and evaluation recipes. We can see that the models could be further boosted, suggesting the compatibility of structural and objective designs.

\begin{table}[htb]
\caption{\textbf{Combinations of MoTo and TokenProp.} }
\label{table:combine}
\centering
\resizebox{\linewidth}{!}{
\begin{tabular}{c|cc|c}
\Xhline{1.1pt}
Model Architecture & w  MoTO & w  TokenProp & Accuracy \\ \hline
DeiT-S             & -       & -            &     79.8     \\
\textbf{Ours}                   &    $\checkmark$     &      $\checkmark$        &    \textbf{82.6}      \\ \hline
T2T-ViT-14         & -       & -            &    81.5      \\
\textbf{Ours}                   &   $\checkmark$      &       $\checkmark$       &   \textbf{82.8}       \\ \hline
Swin-T         & -       & -            &    81.2      \\
\textbf{Ours}                   &     $\checkmark$    &        $\checkmark$      &   \textbf{82.9}  \\
\Xhline{1.1pt}
\end{tabular}}%
\end{table}%

\subsection{Improvements on Data Efficiency}

\begin{table}[htb]
\caption{\textbf{Performance when the model is trained under the resource-limited setting.} The baseline structure uses Swin-T~\cite{liu2021swin} architecture. Each row represents the results trained under a certain portion (percentage) of the original ImageNet-1k training data, for both the baseline and our refined counterpart. Top-1 Accuracy denotes the validation accuracy on original ImageNet validation set.}
\centering
\resizebox{\linewidth}{!}{
\begin{tabular}{c|cccc}
\Xhline{1.1pt}
\multirow{2}{*}{Training Dataset}     & \multirow{2}{*}{Percentage (\%)} & \multicolumn{3}{c}{Val Top-1 Accuracy} \\ \cline{3-5} 
                             &                             & Baseline   &  w Ours   & $\Delta$   \\ \Xhline{1.1pt}
\multirow{4}{*}{ImageNet-1k} & 50                          &     73.7      &     \textbf{75.1}     &    +1.4\%        \\
                             & 40                          &     71.6     &     \textbf{73.2}     &   +1.6\%         \\
                             & 20                          &     61.2      &     \textbf{63.9}     &       +2.7\%     \\
                             & 10                          &    43.5      &    \textbf{46.5}      &     +2.9\%       \\ \Xhline{1.1pt}
\end{tabular}}%
\label{table:data}
\end{table}%

\begin{table*}[htb]
\caption{\textbf{Downstream performance of semantic segmentation on ADE20K dataset}. We modify the tokenizer of DeiT-S and Swin-T with our proposed modules and denote it in bold font. The reported mIoU exploits multi-scale and flip testing.}
\centering
\resizebox{0.95\linewidth}{!}{
\begin{tabular}{c|cccc|cc|c}
\Xhline{1.1pt}
\multirow{2}{*}{Method}                   & \multirow{2}{*}{Backbone}  &  \multicolumn{2}{c}{Module} &   \multirow{2}{*}{Pre-trained}      & \multirow{2}{*}{Crop Size} & \multirow{2}{*}{LR Schedule} & \multirow{2}{*}{mIoU}        \\ \cline{3-4} 

 &  & MoTo  & TokenProp  &   &   &   &  \\ \hline
OCRNet~\cite{yuan2020object}                   & HRNet-w48  &   &   & ImageNet-1k         & 512 $\times$ 512        & 150K        & 45.7      \\ \hline
\multirow{5}{*}{UperNet}        & DeiT-S     &  &    & ImageNet-1k    & 512 $\times$ 512        & 160K        & 44.0      \\ 
                   & DeiT-S w Frozen~\cite{chen2021empirical}  &    &   & ImageNet-1k       & 512 $\times$ 512        & 160K        & 42.9      \\ 
                   & DeiT-S     & $\checkmark$  &     & ImageNet-1k       & 512 $\times$ 512        & 160K        & 44.5      \\ 
                   & DeiT-S &   &    $\checkmark$    & ImageNet-1k       & 512 $\times$ 512        & 160K        & 44.3      \\ 
                   & \textbf{DeiT-S}    &  $\checkmark$ &  $\checkmark$  & ImageNet-1k       & 512 $\times$ 512        & 160K        & \textbf{44.7}      \\  \hline
\multirow{4}{*}{UperNet} & Swin-T  &    &     & ImageNet-1k          & 512 $\times$ 512       & 160K        & 45.8          \\
                         & Swin-T & $\checkmark$    &    & ImageNet-1k  & 512 $\times$ 512       & 160K        & 46.3 \\ 
                         & Swin-T & &   $\checkmark$ & ImageNet-1k  & 512 $\times$ 512       & 160K        & 46.4 \\ 
                         & \textbf{Swin-T} & $\checkmark$ & $\checkmark$  & ImageNet-1k  & 512 $\times$ 512       & 160K        & \textbf{46.7} \\  \hline
\multirow{4}{*}{UperNet} & Swin-S   &   &    & ImageNet-1k          & 512 $\times$ 512       & 160K        & 49.1          \\
                         & Swin-S & $\checkmark$    &    & ImageNet-1k  & 512 $\times$ 512       & 160K        & 49.4 \\                          
                         & Swin-S &     &  $\checkmark$  & ImageNet-1k  & 512 $\times$ 512       & 160K        & 49.4 \\                          
                         & \textbf{Swin-S}  &  $\checkmark$ &  $\checkmark$  & ImageNet-1k  & 512 $\times$ 512       & 160K        & \textbf{49.6} \\                          \Xhline{1.1pt}
\end{tabular}}%
\label{exp:ade}
\end{table*}

The data inefficiency has been a major problem of vision transformers. Since the original training regime of ViT~\cite{dosovitskiy2021an} contains hundreds of millions of training images, efforts have been made to improve its data efficiency that allows training feasibility. Therefore, we investigate whether our proposed strategies benefit the model from better utilization of training data. We incorporate both our strategies to another dominating transformer of Swin transformer~\cite{liu2021swin}, which has already demonstrated strong capability of data exploitation. 

We follow the original training hyper-parameters in~\cite{liu2021swin}, while only leverage a limited portion of ImageNet-1k training set and 100 epochs for training. We train the baseline model under a specific amount of training data, as well as the boosted version with our modules. The results under this resource-limited setting are reported in Table~\ref{table:data}, where a consistent boost is observed under each training portion. The results show that our refinement strategies on vision tokenizer are able to improve the data efficiency, which are not specially designed for this purpose. Meanwhile, better preservation of information accessibility in the extracted tokens builds up their connection and helps the transformer capture more valuable content.

\begin{table*}[htb]
\caption{\textbf{Downstream performance of object detection and instance segmentation on COCO 2017.} We modify the tokenizer of PVT and Swin-T with our proposed modules and denote it in bold font. The backbones are pre-trained on ImageNet-1k dataset.}
\centering
\resizebox{0.8\linewidth}{!}{
\begin{tabular}{c|cc|c|cc}
\Xhline{1.1pt}
Framework                              & Backbone              & Pre-trained & LR Schedule & \begin{tabular}[c]{@{}c@{}}Box\\ mAP\end{tabular} & \begin{tabular}[c]{@{}c@{}}Mask\\ mAP\end{tabular} \\ \Xhline{1.1pt}
\multirow{4}{*}{RetineNet~\cite{lin2017focal}}         & PVTv2-b1              & ImageNet-1k & 1x                         & 41.2                      & -            \\
                & \textbf{PVTv2-b1 w MoTo}          & ImageNet-1k & 1x                & 41.8    & -                          \\ 
                & \textbf{PVTv2-b1 w TokenProp}          & ImageNet-1k & 1x                & 41.5    & -                          \\ 
                & \textbf{PVTv2-b1 w Both}          & ImageNet-1k & 1x                & \textbf{42.0}    & -                          \\ \hline
\multirow{4}{*}{Mask R-CNN~\cite{he2017mask}}         & PVTv2-b1              & ImageNet-1k & 1x                & 41.8          & 38.8        \\
        & \textbf{PVTv2-b1 w MoTo}          & ImageNet-1k & 1x               & 42.4             & 39.1              \\ 
        & \textbf{PVTv2-b1 w TokenProp}          & ImageNet-1k & 1x               & 42.3             & 39.4              \\ 
        & \textbf{PVTv2-b1 w Both}          & ImageNet-1k & 1x               & \textbf{42.9}             & \textbf{39.4}              \\ \hline
\multirow{4}{*}{Mask R-CNN~\cite{he2017mask}}       & Swin-T                & ImageNet-1k & 1x                   & 43.7              & 39.8     \\
    & \textbf{Swin-T w MoTo}    & ImageNet-1k & 1x        & 44.1         & 40.1                                     \\ 
    & \textbf{Swin-T w TokenProp}    & ImageNet-1k & 1x        & 44.2         & 40.0                                      \\ 
    & \textbf{Swin-T w Both}    & ImageNet-1k & 1x        & \textbf{44.6}         & \textbf{40.4}                                      \\ \hline

\multirow{4}{*}{Cascade Mask R-CNN~\cite{cai2018cascade}} & Swin-T             & ImageNet-1k & 1x              & 48.1           & 41.7        \\
                    & \textbf{Swin-T w MoTo} & ImageNet-1k & 1x                 & 48.7           & 42.1                 \\ 
                    & \textbf{Swin-T w TokenProp} & ImageNet-1k & 1x                 & 48.6           & 41.9                 \\ 
                    & \textbf{Swin-T w Both} & ImageNet-1k & 1x                 & \textbf{49.1}           & \textbf{42.4}                 \\ \Xhline{1.1pt}
\end{tabular}}
\label{exp:coco}
\end{table*}

\subsection{\label{supp_sec:downstream}Improvements on Downstream Tasks}

To further validate the effectiveness of the proposed strategies, we perform evaluations with our modules on downstream tasks including semantic segmentation and object detection. For these tasks, we utilize DeiT-S~\cite{touvron2020training}, PVT~\cite{wang2021pyramid}, and Swin-T~\cite{liu2021swin} as the baseline architecture, where our variants include the proposed MoTo and TokenProp.

\subsubsection{Semantic Segmentation}
We evaluate our strategies on semantic segmentation using the ADE20K~\cite{zhou2019semantic} dataset. Details regarding the dataset and hyper-parameters can be found in Appendix~\ref{app:seg}. Similar to the experiments on classification, we validate separately with their respective performance, as well as their ensembles. As shown in Table~\ref{exp:ade}, our method easily improves the model with Swin-T backbone that is already very strong comparing to previous methods. Similar with TokenProp, the randomly-initialized embedding in MoCov3~\cite{chen2021empirical} demonstrates better stability in self-supervised learning. To better understand this with our findings in Section~\ref{sec:structure}, we also implement it in semantic segmentation and compare with TokenProp. The frozen encoding lacks feature expressiveness, which is more crucial for downstream tasks like segmentation.
The results further suggest the importance of proper tokenizer structure, as well as the optimization trade-off between feature quality and accessibility.

\subsubsection{Object Detection and Instance Segmentation}
We also perform evaluations on object detection and instance segmentation using the COCO 2017 dataset. Based on the implementation in MMDetection~\cite{chen2019mmdetection}, we testify the performance on different detection pipelines, including RetinaNet~\cite{lin2017focal}, Mask R-CNN~\cite{he2017mask}, and Cascade Mask R-CNN~\cite{cai2018cascade}. We also compare the performance with different backbones, where the hyper-parameters are provided in Appendix~\ref{app:det}. 

From Table~\ref{exp:coco}, we can see that both our strategies are beneficial for object detection and instance segmentation. We observe a consistent improvement over different backbones and detection pipelines. 

\section{Conclusion and Future Work}
In this work, we demonstrate the important role tokenizer plays in vision transformers. Based on the ``trade-off'' perspective that formulates prominent structural designs, we propose to incorporate better normalization and objective for tokenizers. Extensive experimental results manifest their effectiveness across different transformer structures. 

Our findings further indicate that proper generative supervisory signals help improve discriminative performance. Notably, concurrent works on self-supervised learning with MAE~\cite{he2021masked} and BEiT~\cite{bao2021beit} validates the potential of reconstruction objectives. Generalizing such generative signals to different tasks remains an open problem. 

\ifCLASSOPTIONcaptionsoff
  \newpage
\fi



\bibliographystyle{IEEEtran}
%
\bibliography{egbib}

\appendices

\section{Additional Details}
\subsection{Details on Estimating Conditional Entropy  \label{app:decoder}}
In Section~3 and Section~4, we estimate the empirical mutual information between input images and split tokens with conditional entropy. The metric is computed by training an additional decoder to estimate the reconstruction error. We train the decoder using the Adam optimizer, with the learning rate of 0.001, beta1 0.5, beta2 0.999 on V100 GPUs. As shown in Table~\ref{table:encoder}, we use a simple 3-layer architecture for the decoder. Note that we also adopt this architecture as decoder in TokenProp for simplicity.

\begin{table}[htb]
\caption{\label{table:encoder}Details parameters of decoders. The convolutional layer denotes residual blocks. We also adopt this architecture  for TokenProp. }
\begin{center}\begin{tabular}{c|c|c|c} 
\Xhline{1.2pt}
Layer  & Kernel Size & Output Channel & Output Size \\
\hline
Input w Concat  & - & 196 & 16 \\
\hline
Convolution & 3 & 256 & 16 \\
\hline
Convolution & 3 & 256 & 16 \\
Pixel Shuffle  & - & 64 & 32 \\
\hline
Convolution & 3 & 64 & 32 \\
Pixel Shuffle  & - & 16 & 64 \\
\hline
Convolution & 3 & 3 & 64 \\
\Xhline{1.2pt}
\end{tabular}
\end{center}
\end{table}

\subsection{Details on Comparison across structural designs.}
  \label{app:structure}
Specifically, the intra-token refinement layer consists of kernels with sizes 4$\times$4, 8$\times$8, and 16$\times$16, further concatenating the extracted multi-scale feature and projecting it to the original dimension. Following~\cite{wang2021pvtv2}, the locality strategy enlarges the token window and performs overlapping tokenization with a half-patch step size. Inter-token refinement further adds a self-attention layer before the output. 

For the experiments, we adopt DeiT-S as our baseline architecture and use the consistent training recipes~\cite{touvron2020training} for 300 epochs on ImageNet~\cite{deng2009imagenet} training set for supervised classification. Strictly following MoCov3~\cite{chen2021empirical}, we also conduct unsupervised pre-training and use the linear probing accuracy to test the performance. Semantic segmentation is evaluated consistently using UperNet~\cite{xiao2018unified} backbone and the hyper-parameters in Appendix~\ref{app:seg}.

\section{Implementations and Hyper-parameters}

\subsection{Linear Classification with Self-supervised Training}

In Section~4, we study the performance of linear classification with self-supervised pre-training as in MoCov3~\cite{chen2021empirical}. Here we provide additional details on running these experiments.

\noindent \textbf{Self-supervised Pre-training.~} Following the practice in their official implementation~\cite{code2021moco}, we faithfully implement our variants with DeiT-S backbone. By default we use AdamW~\cite{loshchilov2018fixing}, a batch size of 4096, and a warmup schedule for 10k steps. The learning rate and weight decay is also swept for multiple runs. We also follow the consistent architectures and implementations of the MLP heads and contrastive loss in MoCov3~\cite{chen2021empirical} for fair comparisons.

\noindent \textbf{Linear Probing.~} As a common practice in estimating the representation quality, we evaluate the trained models with linear probing. After the self-supervised pre-training completes, we remove the MLP layers and train an additional linear classification based on the obtained frozen features. We adopt the SGD optimizer with a batch size of 4096. Similarly, the learning rate is swept by multiple runs for better performance.



\subsection{Supervised Classification on ImageNet}
As we've mentioned in Section~5.2, we integrate our proposed module to various state-of-the-art models, including DeiT~\cite{touvron2020training}, Token-2-Token ViT~\cite{yuan2021tokens}, PVT~\cite{wang2021pyramid}, and Swin Transformer~\cite{liu2021swin}. As we don't intend to propose a new training pipeline in this work, we adopt the common practice in training vision transformers. Since these methods both contain distinctive designs and different pipelines for training, we adopt their original training recipes and keep the training parameters consistent with the provided ones, includign data augmentations, batch sizes, and optimizers. As DeiT~\cite{touvron2020training} and Swin serve as two major backbones in this work, we provide the ingredients and hyper-parameters for training in Table~\ref{table:param_cls}.

\begin{table}[htb]
\caption{\label{table:param_cls}Ingredients and hyper-parameters for training DeiT and Swin. For training Swin-S, we use a smaller batch size and a scaled learning rate. EMA represents Exponential Moving Average.}
\begin{center}\begin{tabular}{c|c|c} 
\Xhline{1.2pt}
Methods     &   DeiT-S    &   Swin-T/S    \\ \hline
Training Epochs &   300     &       300  \\  \hline
Batch Size &   1024     &       512  \\
Optimizer &   AdamW     &       AdamW  \\
Learning rate &   0.001     &       0.0005  \\
Weight decay    &   0.05    &       0.05    \\
Warmup epochs   &   5       &       20      \\ \hline
EMA     &   $\checkmark$    &          \\
Stochastic depth    &     $\checkmark$      &   $\checkmark$     \\
Repeated augmentation    &     $\checkmark$      &      \\
Gradient Clip           &      &    $\checkmark$    \\  \hline
Rand Augment        &   $\checkmark$   &   $\checkmark$   \\
Mixup Prob      &     0.8      &    0.8       \\
Cutmix Prob      &     1.0      &    1.0      \\
Erasing Prob      &     0.25      &    0.25       \\
\Xhline{1.2pt}
\end{tabular}
\end{center}
\end{table}

\subsection{\label{app:seg}Semantic Segmentation on ADE20K}
ADE20K dataset consists of 150 different semantic categories, where there are 20210 training images, 2000 validation images, and 3000 testing images. We follow the consistent training recipes in~\cite{liu2021swin} and use UperNet~\cite{xiao2018unified} in MMsegmentation~\cite{mmseg2020}. The models are trained for 160K iterations with the AdamW optimizer, where the initial learning rate and weight decay 
are set to $6 \times 10^{-5}$ and $0.001$. During inference, we also follow the multi-scale testing strategy in~\cite{liu2021swin}, where [0.5, 0.75, 1.0, 1.25, 1.5, 1.75] $\times$ of the training resolution is employed.

\subsection{\label{app:det}Object Detection and Instance Segmentation}
The dataset consists of 118K training images, 5K validation images, and 20K test-dev images. Here we exploit three widely-used detection pipelines, RetinaNet~\cite{lin2017focal}, Mask R-CNN~\cite{he2017mask}, and Cascade Mask R-CNN~\cite{cai2018cascade}. Following the settings and hyper-parameters in~\cite{liu2021swin}, we use AdamW optimizer with an initial learning rate of $10^{-4}$ and a weight decay of 0.05. The training process includes multi-scale training, a batch size of 16, and 1x(12 epochs) training schedule. We re-implement both the baseline models and our variants using their open-source script in MMdetection~\cite{chen2019mmdetection}.

\end{document}